\documentclass[10pt,twocolumn,letterpaper]{article}
\usepackage{multirow}
\usepackage{array}
\usepackage{amsmath}
\usepackage{empheq}
\usepackage{pifont}
\usepackage[table]{xcolor}
\usepackage{cvpr}              
\definecolor{cvprblue}{rgb}{0.21,0.49,0.74}
\usepackage[pagebackref,breaklinks,colorlinks,allcolors=cvprblue]{hyperref}

\def\paperID{45582} 
\def\confName{CVPR}
\def\confYear{2026}

\title{Dual-Path Hyperprior Informed Deep Unfolding Network \\ for Image Compressive Sensing}

\author{Tianyi Lu, Wenxue Cui, Shaohui Liu\\
Harbin Institute of Technology \\
}

\begin{document}
\maketitle
\begin{abstract}
\textit{Recent Deep Unfolding Networks (DUNs) have significantly advanced Compressive Sensing (CS) by integrating iterative optimization with deep networks. However, existing DUNs still suffer from two challenges: 1) Reliance on a single measurement stream, which limits effective information interaction across distinct measurement subsets. 2) Uniform processing of all image regions, which overlooks varying reconstruction difficulties induced by diverse textures.  
To address these limitations, a novel Dual-Path Hyperprior Informed Deep Unfolding Network (DPH-DUN) is proposed, which partitions measurements into double subsets to enable hyperprior-guided reconstruction via a dual-path architecture. In the Deep Hyperprior Learning branch, a series of lightweight neural modules are designed to efficiently generate hyperprior knowledge of different domains, enabling collaborative guidance for the CS reconstruction. In the Hyperprior Informed Reconstruction branch, a deep unfolding framework with hyperprior guidance is constructed to iteratively refine reconstruction. Specifically, \textbf{i)} in the gradient descent step, a Hyperprior Informed Step Size  Generation network is designed to dynamically generate spatially varying step maps, enabling adaptive fine-grained gradient updates. \textbf{ii)} In the proximal mapping step, two well-designed hyperprior informed attention mechanisms are introduced to dynamically focus on challenging regions via gradient-based hard and soft attentions, facilitating CS reconstruction accuracy. Extensive experiments demonstrate that the proposed DPH-DUN outperforms existing CS methods.}

\end{abstract}
\section{Introduction}
\label{sec:intro}


Compressive Sensing (CS) \cite{donoho2006compressed,candes2008introduction} provides an efficient framework for acquiring and reconstructing sparse or compressible signals from far fewer measurements than required by the Shannon–Nyquist theorem \cite{liutkus2014imaging}. By leveraging the intrinsic redundancy of natural signals, CS reduces acquisition and storage costs while preserving reconstruction fidelity, enabling applications in single-pixel imaging \cite{duarte2008single,shin2016single}, hyperspectral imaging \cite{zhang2016locally,zhang2024progressive}, and MRI \cite{lustig2007sparse,9763318}.

\begin{figure}[t]
  \centering
   \includegraphics[width=\linewidth]{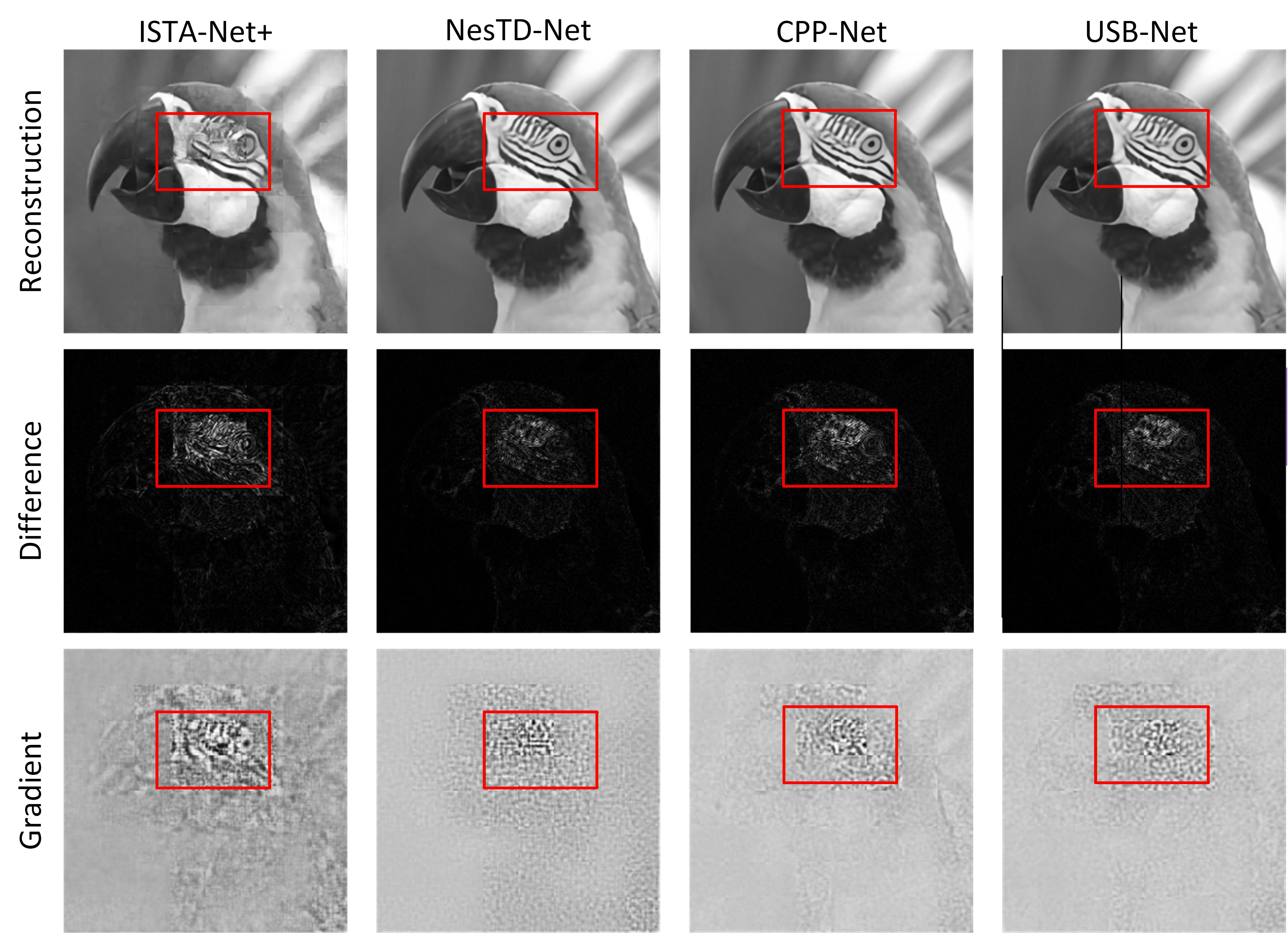}
   \caption{Reconstruction results (top), per-pixel difference maps (middle), and gradient maps (bottom). Larger reconstruction errors are concentrated in regions with high gradients and complex textures, highlighting their inherent reconstruction difficulty.}
   \label{d}
   \vspace{-0.10in}
\end{figure}

The CS problem can be formulated as recovering an unknown signal $\mathbf{x}\in\mathbb{R}^N$ from a set of linear measurements $\mathbf{y}=\boldsymbol{\Phi} \mathbf{x}$, where $\boldsymbol{\Phi}\in\mathbb{R}^{M\times N} (M \ll N)$ denotes the measurement matrix. Since the system is underdetermined, prior knowledge about signal sparsity or structural regularity must be incorporated to make the reconstruction feasible. Accordingly, this process is typically modeled as an optimization problem:
\begin{equation}
\hat{\mathbf{x}} = \arg\min_{\mathbf{x}} \frac{1}{2} \|\boldsymbol{\Phi} \mathbf{x} - \mathbf{y} \|_2^2 + \lambda \mathcal{R}(\mathbf{x}),
\label{e1}
\end{equation}
where the first term $\frac{1}{2} \|\boldsymbol{\Phi} \mathbf{x} - \mathbf{y} \|_2^2$ enforces data fidelity and $\mathcal{R}(\mathbf{x})$ denotes the regularization term incorporating prior knowledge. Designing an appropriate prior remains a core challenge in this ill-posed problem. Traditional approaches exploit sparsity in natural images via transform domains (such as DCT \cite{zhao2014image} or wavelets \cite{akccakaya2010compressed}), and enforce it through $\ell_1$-minimization \cite{usman2018compressive}, Bayesian inference \cite{ji2008bayesian}, or greedy algorithms \cite{lee2016sparse}. Advanced priors include total variation \cite{el2019image,needell2013near}, low-rank constraints \cite{dong2014compressive,majumdar2015improving}, and non-local self-similarity \cite{cui2021image,li2023nonconvex}, yet they involve heavy iterative computation, limiting efficiency and flexibility.


\begin{figure*}[!t]
	\centering
	\includegraphics[width=6.5in]{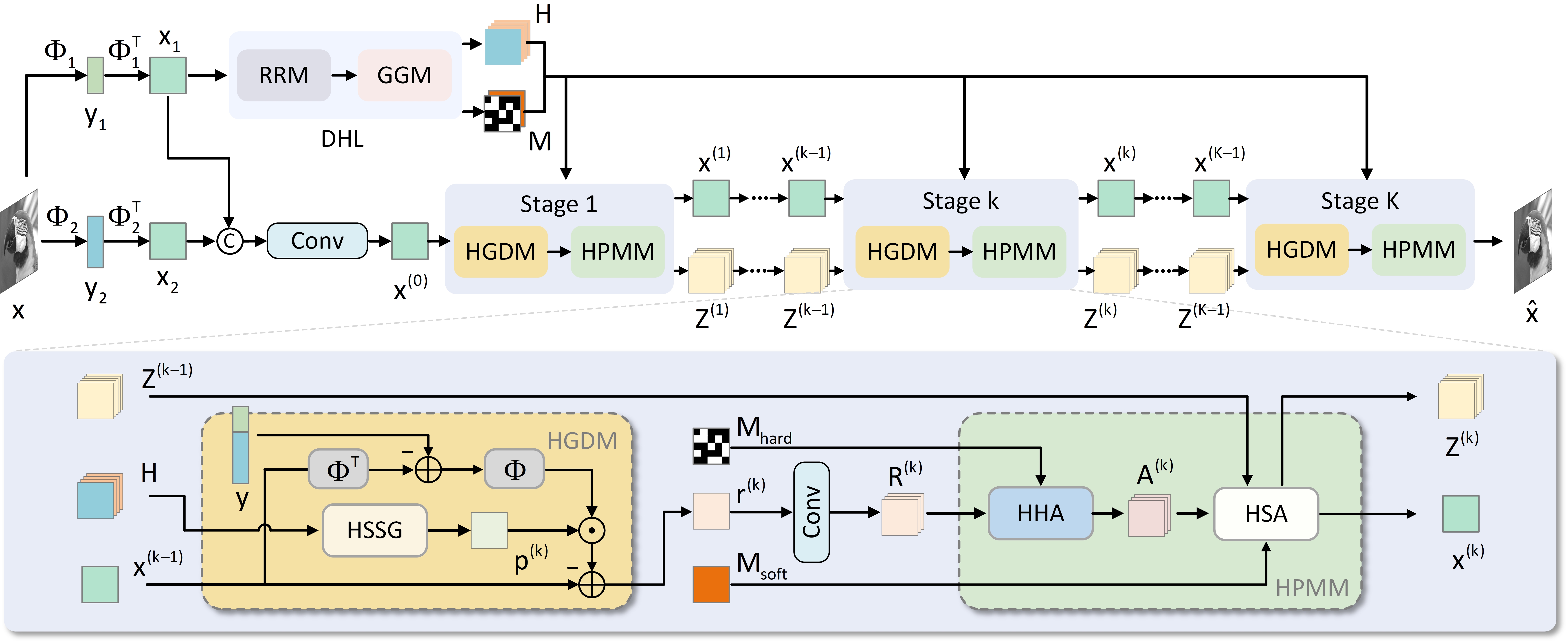}
	\caption{Overview of the proposed DPH-DUN. The upper part shows the dual-path unfolding framework, which consists: the Deep Hyperprior Learning (DHL) branch (top) generates the signal- and gradient-based hyperpriors, and the Hyperprior Informed Reconstruction (HIR) branch (bottom) performs iterative reconstruction through multiple stages consisting of the Hyperprior-Guided Descent Module (HGDM) and the Hyperprior Prior Mapping Module (HPMM). The lower part details the internal structure of each stage.} 
 \vspace{-0.05in}
	\label{fig1}
\end{figure*} 

With the rapid advancement of neural networks, many deep learning-based methods have emerged for CS reconstruction. These methods can be roughly divided into two categories based on their inferential interpretability: \textbf{1) Deep black-box CS methods}  \cite{kulkarni2016reconnet, 9635679, huang2024csdet} directly learn image reconstruction from measurements in an end-to-end manner, without explicit modeling of the underlying reconstruction process. 
By automatically learning informative features, they achieve accurate reconstructions while capturing important image structures. However, due to their black-box nature, these models lack explicit interpretability, making it inherently difficult to examine or regulate their reconstruction behavior. \textbf{2) Deep unfolding CS methods} \cite{song2023DPCDUN, wang2024ufc, 10460452} integrate the mechanisms of traditional iterative solvers into deep network architectures. By unfolding these algorithms into multiple stages, they retain theoretical interpretability while exploiting the feature learning ability of deep networks. 
Compared to deep black-box methods, this hybrid approach provides clearer interpretability, often achieves more accurate reconstructions.


However, existing deep unfolding CS methods still face the challenges in the following two aspects: \textbf{a) Single measurement dependence:} the use of a single measurement stream constrains the information flow and prevents complementary cues from being effectively exploited across measurement subsets, thereby limiting reconstruction adaptivity. \textbf{b) Uniform regional optimization:} the uniform optimization across spatial regions is ineffective in handling structurally complex areas, causing reconstruction errors to concentrate in high-gradient regions (Fig.~\ref{d}) and compromising structural fidelity. 

To address the above challenges, a novel Dual-Path Hyperprior Informed Deep Unfolding Network (DPH-DUN) is proposed, which leverages dual measurement streams and hyperprior knowledge for adaptive CS reconstruction. \textbf{In the Deep Hyperprior Learning (DHL) branch}, lightweight neural modules extract signal- and gradient-based hyperpriors from one measurement subset, providing collaborative guidance for subsequent optimization. \textbf{In the Hyperprior Informed Reconstruction (HIR) branch}, the Proximal Gradient Descent (PGD) process is unrolled into interpretable stages, each consisting of two core modules: \textbf{1)} the Hyperprior Informed Gradient Descent Module (HGDM) employs a Hyperprior Informed Step-Size Generation (HSSG) network to dynamically generate spatially varying step maps, enabling adaptive and fine-grained gradient updates. \textbf{2)} The Hyperprior Informed Proximal Mapping Module (HPMM) applies gradient-based hard and soft attentions to focus on challenging regions, enhancing reconstruction accuracy and structural fidelity.

In summary, our main contributions are as follows:
\begin{enumerate}
    \item{A novel Dual-Path Hyperprior Informed Deep Unfolding CS Network (DPH-DUN) is proposed, which elaborately partitions measurements into two subsets to facilitate hyperprior-guided reconstruction through a dual-path architecture.}
    \item{ In the Deep Hyperprior Learning branch, a series of lightweight neural modules are designed to efficiently generate hyperprior knowledge of different domains, enabling collaborative guidance for the CS reconstruction.}
    \item{In the Hyperprior Informed Reconstruction branch, inspired by the proximal gradient descent algorithm, a deep unfolding framework with hyperprior guidance is constructed to iteratively refine reconstruction.}
    \item{For the gradient descent of the unfolded framework, a Hyperprior Informed Step-size Generation network is designed to dynamically generate spatially varying step maps, enabling adaptive fine-grained gradient updates. For the proximal mapping, two well-designed hyperprior informed attention mechanisms are introduced to focus on challenging regions via gradient-based soft and hard attentions.}
    
\end{enumerate}
\section{Related Work}
\subsection{Deep Black-box CS Methods}
Deep black-box CS methods aim to directly learn the mapping from compressive measurements to the original image without explicitly modeling the sampling or reconstruction process. Early methods  \cite{mousavi2017learning, lohit2018convolutional, xie2017adaptive} often adopt block-based strategies, where the image is divided into small blocks that are independently sampled and reconstructed, then aggregated to form the full image. Notable examples include ReconNet \cite{kulkarni2016reconnet}, which employs fully connected and convolutional layers on each block to learn sampling and reconstruction. Although integrating denoisers such as BM3D \cite{dabov2007image} reduces some artifacts, block-wise processing inherently introduces blocking artifacts and limits reconstruction fidelity.

To address these issues, later deep black-box methods \cite{2019Image, 2019Scalable, 9159912} adopt full-image reconstruction. CSNet \cite{2019Image}, for instance, converts all measurements into a coarse image estimate, which is then refined through a deep convolutional network to improve structural consistency. NL-CSNet \cite{9635679} further incorporates non-local operations to capture long-range dependencies in both measurement and feature domains. More recent transformer-based approaches, such as TCS-Net \cite{10049603} and MTC-CSNet \cite{shen2024mtc}, leverage hierarchical or dual-branch designs to model global context and inter-pixel dependencies. By processing the full image context, these methods reduce blocking artifacts and produce reconstructions with more coherent structures compared to block-based methods.

\subsection{Deep Unfolding CS Methods}
Deep unfolding CS methods integrate iterative optimization algorithms into trainable deep networks, combining the interpretability of optimization-based methods with the representation power of deep networks. Early methods \cite{9298950,9854112,8550778}, such as ISTA-Net \cite{zhang2018ista}, unfold the traditional ISTA algorithm into multi-stage networks, where each stage reconstructs the full image and iteratively refines the previous estimate via learned proximal operators or deblocking modules. AMP-Net \cite{9298950} and FSOINet \cite{2022FSOINet} follow similar strategies, transmitting the reconstructed image across stages. While these methods are interpretable, transmitting only the image estimate between stages may cause information loss and limit reconstruction quality.

Recent deep unfolding CS methods \cite{wang2024ufc,song2023DPCDUN,10655712} enhance feature propagation across stages to preserve fine details and improve reconstruction fidelity. For example, DUN-CSNet \cite{10269792} integrates intermediate features into the gradient descent and proximal mapping modules, strengthening the modeling of complex textures. D3C2-Net \cite{li2024d} introduces a dual-domain coding mechanism to enable efficient cross-stage feature communication. Similarly, CPP-Net \cite{10655712} employs multi-scale feature extraction and stage-wise fusion to maintain subtle structures. These strategies collectively demonstrate that richer inter-stage feature communication leads to higher-quality CS reconstruction.

Compared to deep black-box CS methods, deep unfolding CS methods generally provide a more principled framework and better interpretability by explicitly integrating optimization procedures into the network design. However, existing CS methods often treat all image regions or features uniformly and perform global operations, which may result in uneven reconstruction quality across complex and simple textures and lead to redundant computation in less informative areas, limiting efficiency and scalability.

\section{Method}
\label{sec:formatting}
\subsection{Overview of DPH-DUN}
The overview of the proposed DPH-DUN is illustrated in Fig.~\ref{fig1}. Given an image $\mathbf{x}$, two different sampling branches acquire block-based compressive measurements:
\begin{equation}
\mathbf{y} = \left[ \begin{array}{c}
\mathbf{y}_1 \\
\mathbf{y}_2
\end{array} \right] = \left[ \begin{array}{c}
\boldsymbol{\Phi}_1 \\
\boldsymbol{\Phi}_2
\end{array} \right] \mathbf{x} = \boldsymbol{\Phi} \mathbf{x}
\end{equation}
where $\boldsymbol{\Phi}_1$ and $\boldsymbol{\Phi}_2$ are sampling matrices of two branches. 

After acquiring two measurements, the network employs a dual-path architecture. One branch extracts hyperpriors from one measurement subset to encode content and structural characteristics, while the other branch leverages these hyperpriors to guide the reconstruction of the target image. 
This dual-path operation is formulated as:
\begin{empheq}[left=\empheqlbrace]{align}
	&\mathbf{H},\mathbf{M}=\text{DHL}\left(\mathbf{y}_1,\boldsymbol{\Phi}_1 \right) \label{1}\\
	&\hat{\mathbf{x}}=\text{HIR}\left(\mathbf{y},\boldsymbol{\Phi},\mathbf{H},\mathbf{M}\right) \label{2}
\end{empheq}
where $\mathbf{H}$ and $\mathbf{M}$ represent the signal- and the gradient-based hyperpriors, respectively.



The reconstruction in Eq.~\ref{2} is implemented via a stage-wise PGD procedure to solve the CS problem in Eq.~\ref{e1}, unrolled over $K$ stages:
\begin{empheq}[left=\empheqlbrace]{align}
	&\mathbf{r}^{\left(k\right)}=\mathbf{x}^{\left(k-1\right)}-\mathcal{P}^{\left(k\right)}(\mathbf{H}) \Phi^{\text{T}} (\Phi \mathbf{x}^{(k-1)} - \mathbf{y}) \label{g}\\
    &\mathbf{x}^{\left(k\right)}=\mathcal{H}_{\lambda,\mathcal{R}}^{\left(k\right)}\left(\mathbf{r}^{\left(k\right)}, \mathbf{M}\right) \label{p}
\end{empheq}
where 
$\mathbf{x}^{(0)}$ denotes the initial reconstruction obtained from two measurements, as detailed in Sec.~\ref{irm}. 
Corresponding to Eqs.~\ref{1} and~\ref{2}, DPH-DUN adopts a dual-branch design, where two branches cooperate to progressively reconstruct the target image. In the first branch, the hyperpriors are extracted to provide guidance for subsequent optimization. In the second branch, the optimization process outlined in Eqs.~\ref{g} and \ref{p} involves two critical modules at each stage: the Gradient Descent Module (GDM) and the Proximal Mapping Module (PMM). Through $K$ stages, the framework progressively improves structural fidelity and texture, producing the final reconstruction $\hat{\mathbf{x}} = \mathbf{x}^{(K)}$.

Specifically, the Deep Hyperprior Learning (DHL) branch generates two different hyperpriors using a series of lightweight neural modules: the signal-based hyperpriors that provide content-aware information and the gradient-based hyperpriors that offer structure-aware guidance. Then the Hyperprior Informed Reconstruction (HIR) branch is progressively refined by two key components (Fig.~\ref{fig1}): the Hyperprior Informed Gradient Descent Module (HGDM), which generates spatially varying step maps guided by signal-based hyperpriors to adaptively modulate gradient updates, and the Hyperprior Informed Proximal Mapping Module (HPMM), which applies hard and soft attentions guided by gradient-based hyperpriors to focus on challenging regions, improving reconstruction accuracy and preserving fine structures.

\subsection{Initial Reconstruction}
\label{irm}
For each measurement, the coarse estimate is obtained by projecting the block-based measurements back into the image domain through the corresponding sensing matrix:
\begin{equation}
\mathbf{x}_1 = \boldsymbol{\Phi}_1^{\text{T}} \mathbf{y}_1, \quad \mathbf{x}_2 = \boldsymbol{\Phi}_2^{\text{T}} \mathbf{y}_2.
\end{equation}

This operation produces two coarse estimates, $\mathbf{x}_1$ and $\mathbf{x}_2$, each representing a partial recovery from its respective sampling pattern. The two coarse estimates are then fused to form the initial reconstruction $\mathbf{x}^{(0)}$ for iterative refinement: 

\begin{equation}
	\mathbf{x}^{(0)}=\text{Conv}(\text{Concat}(\mathbf{x}_1,\mathbf{x}_2)),
\end{equation}
where $\text{Concat}(\cdot)$ denotes channel-wise concatenation and $\text{Conv}(\cdot)$ ensures that the fused output matches the original image size. 

\begin{figure}[t]
  \centering
   \includegraphics[width=
   0.95\linewidth]{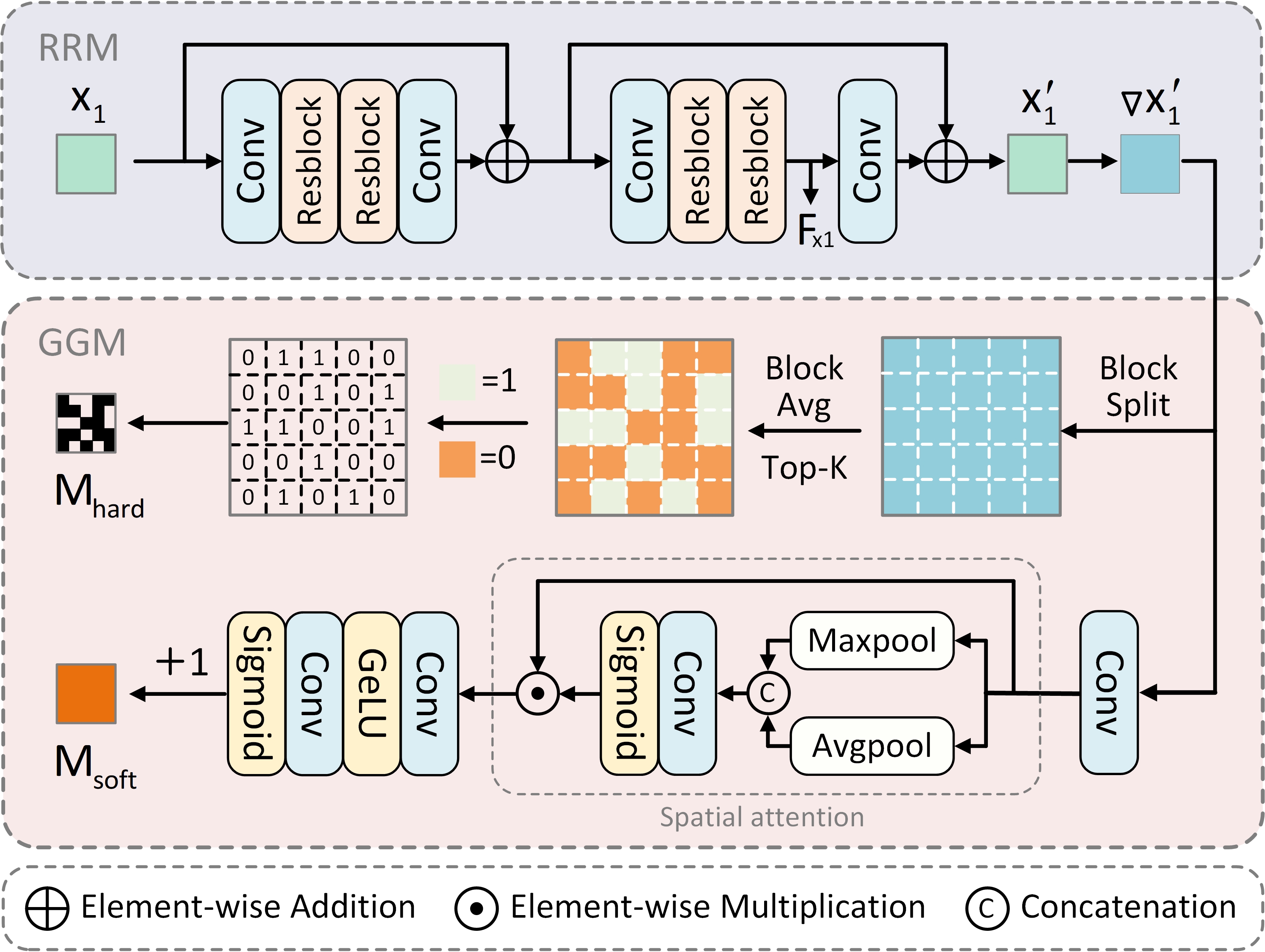}
   \vspace{-0.05in}
   \caption{Details of the Deep Hyperprior Learning (DHL), which consists of a Reconstruction Refinement Module (RRM) and a Guidance Generation Module (GGM).}
   \label{HE}
   \vspace{-0.10in}
\end{figure}

\subsection{Deep Hyperprior Learning}
In the Deep Hyperprior Learning (DHL) branch, illustrated in Fig.~\ref{HE}, two functional modules are designed: the Reconstruction Refinement Module (RRM) and the Guidance Generation Module (GGM). The RRM enhances the coarse estimate and extracts informative representations as signal-based hyperpriors, while the GGM transforms the gradient cues into gradient-based hyperpriors, including a hard structural mask and a soft confidence map. 

\paragraph{Reconstruction Refinement Module.}
Given the coarse estimate $\mathbf{x}_1$, the RRM refines its representation through convolutional and residual blocks to produce an enhanced reconstruction $\mathbf{x}_{1}'$ and intermediate features $\mathbf{F}_{\text{x1}}$. From the enhanced reconstruction, the gradient map is computed as:
\begin{equation}
    \mathrm{\nabla \mathbf{x}_1'}  = \boldsymbol{\Phi}_1^{\text{T}} (\boldsymbol{\Phi}_1 \mathbf{x}_1' - \mathbf{y}_1),
\end{equation}
where both $\mathbf{F}_{\text{x1}}$ and the gradient map $\mathrm{\nabla \mathbf{x}_1'}$ constitute the signal-based hyperpriors $\mathbf{H}$, which encode content dependencies for adaptive descent modulation.

\paragraph{Guidance Generation Module.}
The GGM utilizes the gradient map $\mathrm{\nabla \mathbf{x}_1'}$ to derive gradient-based hyperpriors $\mathbf{M}$ that emphasize structural information. Specifically, it produces two complementary guidance signals: a hard structural mask $\mathbf{M}_{\text{hard}}$ and a soft confidence map $\mathbf{M}_{\text{soft}}$.

The hard structural mask highlights salient regions at the block level. The image is partitioned into non-overlapping blocks $\{\mathcal{B}_b \}_{b=1}^B$, and the mean gradient magnitude in each block is computed as:



\begin{equation}
    g_b = \frac{1}{|\mathcal{B}_b|} \sum_{(i,j) \in \mathcal{B}_b} \mathrm{\nabla \mathbf{x}_1'}(i,j),
\end{equation}
where $g_b$ indicates the average gradient strength. The top-$K$ blocks with the largest $g_b$ values form the binary mask: 
\begin{equation}
    \mathbf{M}_{\text{hard}}(i,j) = 
\begin{cases}
1, &  (i,j) \in \text{Top-K blocks}, \\
0, & \text{otherwise.}
\end{cases}
\end{equation}

To complement the block-level representation, a soft confidence map $\mathbf{M}_{\text{soft}}$ provides the pixel-wise attention. A lightweight spatial attention module enhances the gradient map in the feature domain:
\begin{equation}
    \mathbf{F}_\text{c} = \text{SA}(\text{Conv}_\text{proj}(\mathrm{\nabla \mathbf{x}_1'})),	
\end{equation}
followed by a convolutional mapping to produce the soft confidence map:
\begin{equation}
    \mathbf{M}_{\text{soft}} = 1 + \text{Sigmoid}(\text{Conv}(\text{GeLU}(\text{Conv}(\mathbf{F}_\text{c})))).	
\end{equation}

In this way, DHL branch extracts signal- and gradient-based hyperpriors to comprehensively encode both content and structural information, establishing a unified representation that effectively guides the subsequent reconstruction.

\subsection{Hyperprior Informed Reconstruction}
\subsubsection{Hyperprior Informed Gradient Descent Module}
\paragraph{Overview.}
As defined in Eq.~\ref{g}, the Hyperprior Informed Gradient Descent Module (HGDM) realizes the gradient descent step in the unrolled PGD framework, where the step-size operator $\mathcal{P}^{\left(k\right)}$ is instantiated by the Hyperprior Informed Step-size Generation (HSSG) network. 

Specifically, HSSG predicts a spatially adaptive step-size map $\mathbf{p}^{(k)}$ based on the signal-based hyperprior $\mathbf{H}$ and the stage-wise modulation factor:

\begin{equation}
    \mathbf{p}^{(k)} = \text{HSSG}(\mathbf{H}, \mathbf{m}_\text{stage}^{(k)}),
\end{equation}
where $\mathbf{m}_\text{stage}^{(k)} = \frac{k}{K}$ $(k=1,2,\ldots,K)$ encodes the relative progress of the current iteration and is broadcast to match spatial dimensions before concatenation with other inputs.

\begin{figure}[t]
	\centering
	\includegraphics[width=0.96\linewidth]{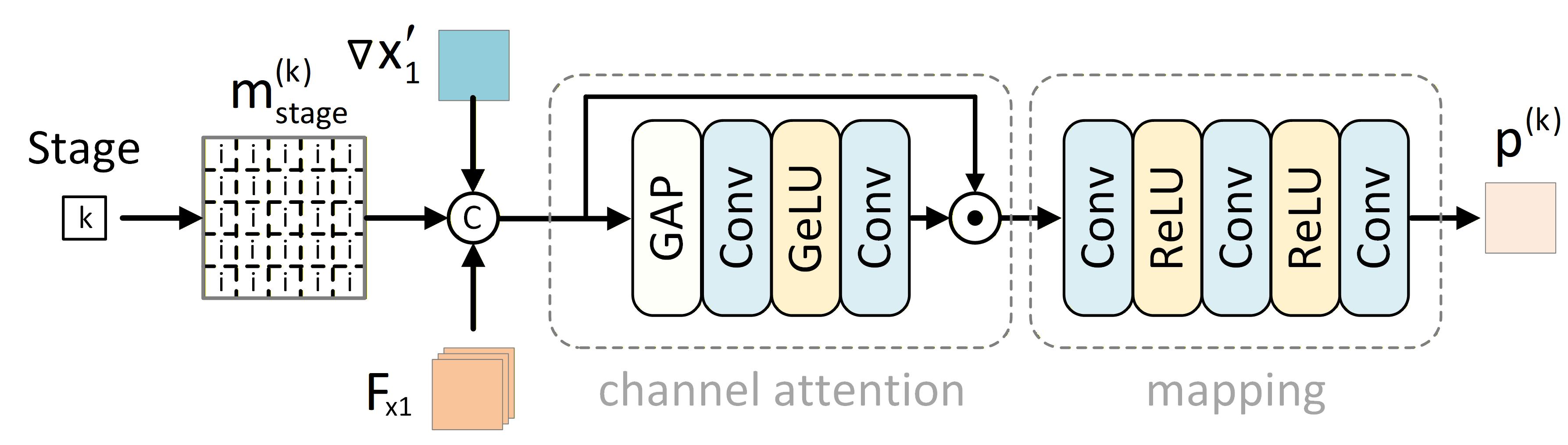}
    \vspace{-0.05in}
	\caption{Details of the Hyperprior Informed Step-size Generation Generator (HSSG) network.} 
	\label{ssg}
    \vspace{-0.05in}
\end{figure}

\paragraph{Hyperprior Informed Step-size Generation.}
As illustrated in Fig.~\ref{ssg}, HSSG network first concatenates the signal-based hyperpriors with the stage-wise modulation factor:
\begin{equation}
    \mathbf{F}_\text{in} = \text{Concat}(\nabla \mathbf{x}_1', \mathbf{F}_\text{x1}, \mathbf{m}_\text{stage}^{(k)}),
\end{equation}
and applies a lightweight channel attention block to adaptively highlight the contributions of these inputs for each channel: 
\begin{equation}
    \mathbf{F}_\text{ca} = \text{Conv}(\text{GeLU}(\text{Conv}(\text{GAP}(\mathbf{F}_\text{in})))) \odot \mathbf{F}_\text{in},
\end{equation}
where $\text{GAP}(\cdot)$ denotes global average pooling, and $\odot$ is element-wise multiplication.

The features are then processed through a stack of convolution and activation layers to produce the step-size map:
\begin{equation}
    \mathbf{p}^{(k)} = \text{Conv}(\text{ReLU}(\text{Conv}(\text{ReLU}(\text{Conv}(\mathbf{F}_\text{ca}))))).
\end{equation}

This design enables HSSG to generate step-size maps that are both spatially adaptive and stage-aware, leveraging hyperprior guidance and iterative stage information, which improves convergence stability and reconstruction fidelity.

\begin{figure}[t]
  \centering
   \includegraphics[width=\linewidth]{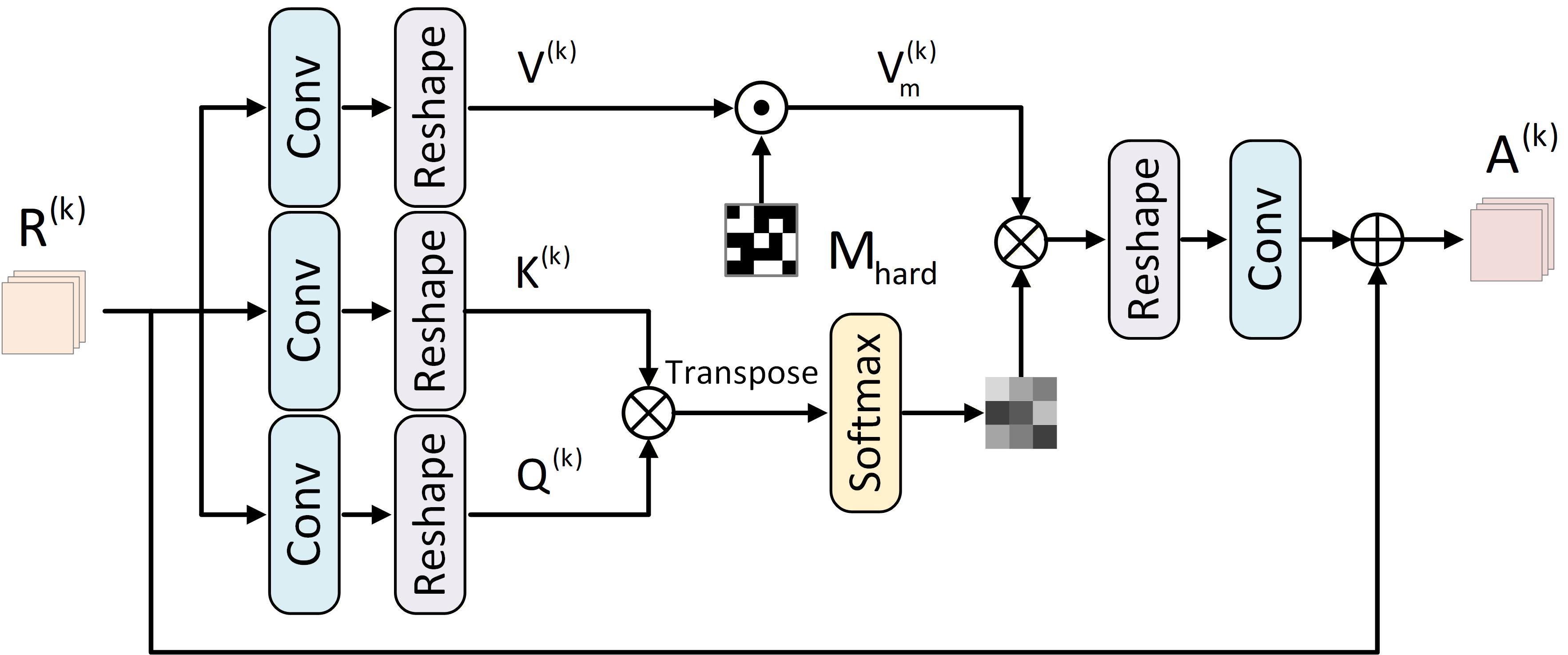}
   \caption{Details of the Hyperprior Informed Hard Attention (HHA). $\otimes$ represents the Matrix Multiplication.}
   \vspace{-0.15in}
   \label{fig:RFA}
\end{figure}

\subsubsection{Hyperprior Informed Proximal Mapping Module}
\paragraph{Overview.}
As shown in Fig.~\ref{fig1}, the Hyperprior Informed Proximal Mapping Module (HPMM) refines the HGDM's output and generates the reconstructed image. 

First, the input $\mathbf{r}^{(k)}$ is projected into the feature domain to obtain the feature representation $\mathbf{R}^{(k)}$, which is then refined by the Hyperprior Informed Hard Attention (HHA) using the hard structural mask $\mathbf{M}_{\text{hard}}$ to emphasize structural details:
\begin{equation}
    \mathbf{A}^{(k)} = \text{HHA}(\mathbf{R}^{(k)}, \mathbf{M}_{\text{hard}}).
\end{equation}

Then, the soft confidence map $\mathbf{M}_{\text{soft}}$ applies an element-wise weighting to $\mathbf{A}^{(k)}$ by the Hyperprior Informed Soft Attention (HSA) at a multi-scale level.
Combined with the previous feature $\mathbf{Z}^{(k-1)}$, HSA produces the refined reconstruction $\mathbf{x}^{(k)}$ and updated feature $\mathbf{Z}^{(k)}$:
\begin{equation}
    (\mathbf{x}^{(k)},\mathbf{Z}^{(k)}) = \text{HSA}(\mathbf{A}^{(k)}, \mathbf{Z}^{(k-1)},\mathbf{M}_{\text{soft}}).
\end{equation}

By combining hard and soft hyperprior informed attentions, HPMM enhances structural fidelity and improves adaptive feature refinement during reconstruction.
\vspace{-0.05in}
\paragraph{Hyperprior Informed Hard Attention.}
To enhance structurally salient regions, we introduce the Hyperprior Informed Hard Attention (HHA, Fig.~\ref{fig:RFA}), guided by the hard structural mask from DHL. By emphasizing gradient-intensive blocks, HHA suppresses background redundancy and reinforces structure-aware information.

Given the input feature $\mathbf{R}^{(k)}$, the query, key, and value are obtained by linear projections:
\begin{equation}
\small
    \mathbf{Q}^{(k)} = \mathbf{W}_\text{Q}\mathbf{\mathbf{R}^{(k)}},
    \mathbf{K}^{(k)} = \mathbf{W}_\text{K}\mathbf{\mathbf{R}^{(k)}},
    \mathbf{V}^{(k)} = \mathbf{W}_\text{V}\mathbf{\mathbf{R}^{(k)}}.
\end{equation}
where $\mathbf{W}_\text{Q}$, $\mathbf{W}_\text{K}$, $\mathbf{W}_\text{V}$ represent the projection weights.
Rather than aggregating information uniformly, HHA selectively enhances value features at structurally important regions indicated by the hard structural mask:
\begin{equation}
    \mathbf{V}_\text{m}^{(k)} = \mathbf{V}^{(k)} \odot {\mathbf{M}}_{\text{hard}},
\end{equation}
where $\odot$ is element-wise multiplication. 
The final feature is then obtained by computing attention over these masked value features:
\begin{equation}
    \mathbf{A}^{(k)} = \text{Softmax}\left(\frac{\mathbf{Q}^{(k)}\mathbf{K}^{(k){\text{T}}}}{\sqrt{d}}\right) \mathbf{V}_\text{m}^{(k)} + \mathbf{R}^{(k)}.
\end{equation}
emphasizing structurally important blocks for subsequent processing.
\vspace{-0.1in} 
\paragraph{Hyperprior Informed Soft Attention.}
\begin{figure}[t]
  \centering
   \includegraphics[width=\linewidth]{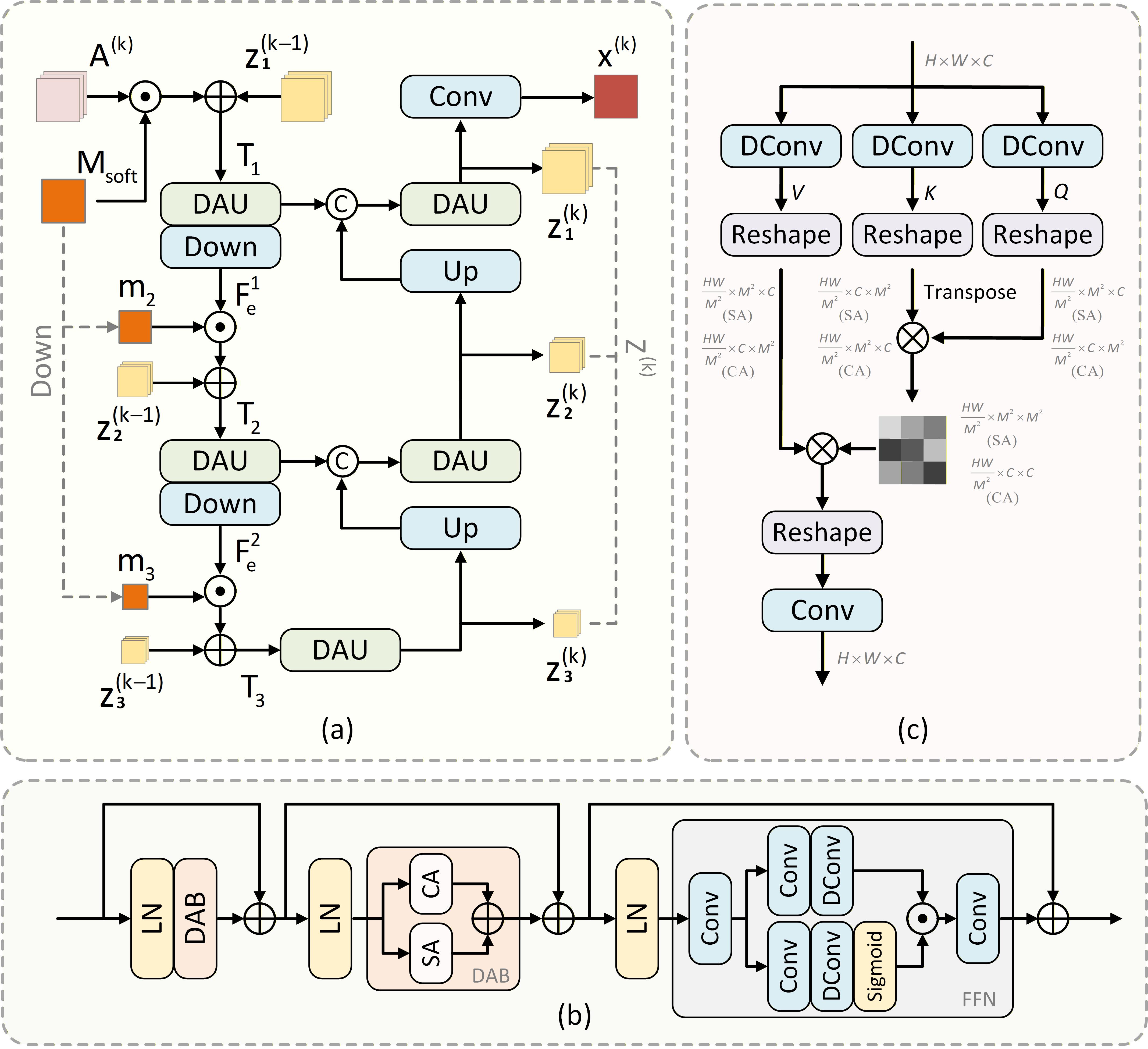}
    \vspace{-0.2in}
   \caption{Details of the Hyperprior Informed Soft Attention (HSA). (a) HSA (b) Dual-Attention Unit (DAU). (c) Spatial attention (SA) and channel attention (CA) in DAB.}
   \vspace{-0.1in}
   \label{fig:MAEM}
\end{figure}

As illustrated in Fig.~\ref{fig:MAEM} (a), the Hyperprior Informed Soft Attention (HSA) refines HHA's output under the guidance of the soft confidence map at each scale. Hierarchical features from the previous stage, $\mathbf{Z}^{(k-1)}=\{\mathbf{Z}_1^{(k-1)},\mathbf{Z}_2^{(k-1)},\mathbf{Z}_3^{(k-1)}\}$, are incorporated to provide cross-iteration cues. 

The confidence map is injected into multiple scales through a modulation–aggregation–refinement process. To align with the resolutions of hierarchical features, the soft confidence map ${\mathbf{M}}_{\text{soft}}$ is resized as:
\begin{equation}
\mathbf{m}_i =
\begin{cases}
\mathbf{M}_\text{soft}, & i=1,\\
\text{Down}(\mathbf{m}_{i-1}), & i=2,3.
\end{cases}
\end{equation}

At each encoder scale, the feature is modulated and fused with the previous stage’s representation:
\begin{equation}
\mathbf{T}_e^{i} = \mathbf{m}_i \odot \mathbf{F}_e^{i-1} + \mathbf{Z}_i^{(k-1)}, \quad (i=1,2,3),
\end{equation}
where $\mathbf{F}_e^{0}$ is initialized by $\mathbf{A}^{(k)}$. The output is processed by the Dual-Attention Unit (DAU, Fig.~\ref{fig:MAEM}(b)), composed of Layer Normalization (LN), Dual-Attention Block (DAB), and Feed-Forward Network (FFN), which adaptively enhances spatial and channel cues. The refined feature is then downsampled to obtain $\mathbf{F}_e^{i}$ for the next scale.

The decoder mirrors the encoder, progressively fusing and refining multi-scale features via DAUs to reconstruct the full-resolution feature, which is finally aggregated through a convolutional layer to produce the output.
\begin{equation}
\mathbf{x}^{(k)} = \text{Conv}(\mathbf{Z}_{1}^{(k)}),
\end{equation}
where $\mathbf{Z}_i^{(k)}$ denotes the decoder output at scale $i$, and $\mathbf{Z}_1^{(k)}$ represents the final full-resolution feature.

 \begin{table*}[ht]
	\centering
	\caption{Average PSNR (dB)/SSIM performance comparisons of CS methods on various datasets at CS ratios $\gamma\in\left \{0.01,0.04,0.10,0.25\right \}$. The best results are highlighted in \textcolor{red}{red}, and the second-best results are highlighted in \textcolor{blue}{blue}.}
    \tiny
    \setlength{\tabcolsep}{2.5pt} 
	\begin{tabular}{c|c|cccccccccccc|cc}
		\toprule
         \multirow{2}{*}{Dataset} & \multirow{2}{*}{$\gamma$} & ISTA-Net+ & AMP-Net & CSNet+ & BCSNet & FOSINet & TCS-Net & DPC-DUN & OCTUF & NesTD-Net & UFC-Net & CPP-Net & USB-Net & DPH-DUN$^*$&DPH-DUN \\ 
         &&(CVPR 2018)&(TIP 2020)&(TIP 2020)&(TMM 2021)&(ICASSP 2022)&(TCI 2023)&(TIP 2023)&(CVPR 2023)&(TIP 2024)&(CVPR 2024)&(CVPR 2024)&(TIP 2025)&(Ours)&(Ours) \\ \midrule
        \multirow{5}{*}{Urban100} 
		& 0.01 & 16.67/0.3734 & 19.62/0.5025 & 19.13/0.4602 & 19.46/0.4795 & 19.87/0.5223 & 19.61/0.4945 & 17.31/0.4216 & 19.88/0.5167 & 20.13/0.5288 & 19.69/0.5041 & 20.55/0.5554 & 20.64/0.5617 & \textcolor{blue}{21.42/0.6108} &\textcolor{red}{21.52/0.6167}\\ 
		& 0.04 & 19.66/0.5370 & 22.81/0.6825 & 21.93/0.6486& 22.43/0.6982&23.69/0.7376&22.93/0.7036&22.36/0.6768&23.68/0.7329&23.94/0.7432&23.37/0.7195&24.66/0.7691&24.73/0.7714& \textcolor{blue}{25.52/0.7954} &\textcolor{red}{25.87/0.8051}\\ 
		& 0.10 & 23.51/0.7201 & 26.04/0.8151 & 24.72/0.7929& 25.25/0.8103&27.53/0.8627&25.87/0.8291&26.96/0.8361&27.79/0.8621&27.80/0.8681&27.55/0.8583&28.49/0.8801&28.58/0.8818&\textcolor{blue}{29.46/0.8941}&\textcolor{red}{29.82/0.8990} \\ 
		& 0.25 & 28.91/0.8834 & 30.89/0.9202 & 28.08/0.8907&29.55/0.9101&32.62/0.9430&30.13/0.9241&32.36/0.9323&32.99/0.9445&33.02/0.9448&32.82/0.9423&33.38/0.9485&33.56/0.9500& \textcolor{blue}{34.13/0.9532}&\textcolor{red}{34.64/0.9557}\\ 
		\cmidrule(r){2-16}
		& Avg. & 22.19/0.6285 & 24.84/0.7301 & 23.47/0.6981& 24.17/0.7245&25.93/0.7664&24.64/0.7378& 24.75/0.7167&26.09/0.7641&26.22/0.7712&25.86/0.7561&26.77/0.7883&26.88/0.7912&\textcolor{blue}{27.63/0.8134} &\textcolor{red}{27.96/0.8191}\\ \midrule
		\multirow{5}{*}{Set11} 
		& 0.01 & 17.45/0.4131 & 20.20/0.5581 & 20.69/0.5238 & 20.81/0.5427 & 21.73/0.5937&21.09/0.5505&18.03/0.4601&21.75/0.5934&21.40/0.5891&21.24/0.5607&22.19/0.6135&22.29/0.6168& \textcolor{blue}{22.82/0.6505} &\textcolor{red}{22.89/0.6532}\\ 
		& 0.04 & 21.56/0.6240 & 25.26/0.7722 & 24.54/0.7445 & 24.90/0.7531 & 26.37/0.8119&25.46/0.7863&24.38/0.7498&26.45/0.8126&26.73/0.8238&25.92/0.7943&27.23/0.8337&27.26/0.8335&\textcolor{blue}{27.57/0.8433} &\textcolor{red}{27.71/0.8463} \\ 
		& 0.10 & 26.49/0.8036 & 29.40/0.8779 & 28.12/0.8664 & 29.36/0.8650 & 30.44/0.9018&29.04/0.8834&29.42/0.8801&30.70/0.9030&30.91/0.9099&30.15/0.8960&31.27/0.9135&31.31/0.9149&\textcolor{blue}{31.74/0.9199}&\textcolor{red}{31.86/0.9222}\\ 
		& 0.25 & 32.44/0.9237 & 34.63/0.9481 & 32.20/0.9337 & 34.20/0.9480 & 35.80/0.9595&33.94/0.9508&34.75/0.9483&36.10/0.9604&36.27/0.9622&35.42/0.9567&36.35/0.9631&36.42/0.9632& \textcolor{blue}{36.70/0.9645}&\textcolor{red}{36.93/0.9654} \\
		\cmidrule(r){2-16}
		& Avg. & 24.49/0.6911 & 27.37/0.7891 & 26.39/0.7671 & 27.32/0.7772 & 28.59/0.8167 &27.38/0.7928&26.65/0.7596& 28.75/0.8174 & 28.83/0.8213& 28.18/0.8019 & 29.26/0.8310 &29.32/0.8321 & \textcolor{blue}{29.71/0.8446} &\textcolor{red}{29.85/0.8468} \\ \midrule
		\multirow{5}{*}{Set14} 
        & 0.01 & 18.22/0.4014 & 21.64/0.5433 & 21.18/0.4987 & 21.45/0.5172 &22.00/0.5538&21.64/0.5219&19.04/0.4551&21.94/0.5500&22.32/0.5600&21.79/0.5324&22.52/0.5694&22.59/0.5712& \textcolor{red}{23.30}\textcolor{blue}{/0.6011} &\textcolor{blue}{23.26}\textcolor{red}{/0.6012}\\ 
		& 0.04 & 22.08/0.5708 & 25.50/0.7007 & 24.57/0.6816 & 24.94/0.7009 &26.08/0.7324&25.25/0.7073&24.32/0.6630&26.04/0.7302&26.31/0.7393&25.67/0.7163&26.54/0.7440&26.55/0.7441& \textcolor{blue}{26.94/0.7559} &\textcolor{red}{27.06/0.7589}\\ 
		& 0.10 & 26.00/0.7289 & 28.77/0.8183 & 27.51/0.8134 & 27.92/0.8199 &29.35/0.8451&28.19/0.8283&28.03/0.7950&29.47/0.8454&29.62/0.8504 & 29.10/0.8363 & 29.93/0.8537 & 29.90/0.8542 & \textcolor{blue}{30.26/0.8596}&\textcolor{red}{30.42/0.8615} \\ 
		& 0.25 & 30.62/0.8700 & 33.21/0.9144 & 31.10/0.9065&32.11/0.9172&34.05/0.9309&32.23/0.9206& 32.78/0.9023&34.18/0.9312&34.33/0.9330&33.81/0.9259&34.41/0.9336&34.43/0.9343& \textcolor{blue}{34.80/0.9360} &\textcolor{red}{34.94/0.9363} \\ 
		\cmidrule(r){2-16}
		& Avg. & 24.23/0.6428 & 27.28/0.7442 &  26.09/0.7251&26.62/0.7388&27.87/0.7656&26.83/0.7445&26.04/0.7039&27.91/0.7642&28.15/0.7707& 27.59/0.7527 &28.35/0.7752& 28.37/0.7759&\textcolor{blue}{28.82/0.7882} &\textcolor{red}{28.92/0.7895}\\ \midrule
		\multirow{5}{*}{General100} 
		& 0.01 & 19.00/0.4700 & 22.71/0.6282 & 21.68/0.5705 & 22.28/0.5877 & 23.27/0.6363 & 22.58/0.5978 & 19.95/0.5363& 23.31/0.6346 & 23.14/0.6165 & 23.08/0.6145 & 24.17/0.6593 & 24.29/0.6620 & \textcolor{blue}{25.09/0.6898} & \textcolor{red}{25.10/0.6910} \\ 
		& 0.04 & 23.76/0.6549 & 26.96/0.7695 & 25.33/0.7403 & 26.02/0.7561 & 28.39/0.8135 & 26.57/0.7712 & 26.61/0.7531& 28.35/0.8122 & 28.58/0.8211 & 27.92/0.7988 & 29.04/0.8284 & 29.10/0.8287 & \textcolor{blue}{29.57/0.8371} & \textcolor{red}{29.76/0.8403} \\ 
		& 0.10 & 28.54/0.8104 & 30.82/0.8722 & 28.69/0.8690 & 29.81/0.8719 & 32.70/0.9085 & 29.90/0.8748 & 31.17/0.8716 & 32.77/0.9084 & 32.85/0.9123 & 32.31/0.9014 & 33.16/0.9159 & 33.27/0.9166 & \textcolor{blue}{33.77/0.9192} &  \textcolor{red}{33.91/0.9205}\\ 
		& 0.25 & 34.32/0.9250 & 36.01/0.9508 & 32.68/0.9348 & 33.52/0.9401 & 38.13/0.9660 & 34.63/0.9504 & 36.50/0.9481 & 38.26/0.9666 & 38.42/0.9670 & 37.75/0.9624 & 38.32/0.9674 & 38.38/0.9681 & \textcolor{blue}{38.85/0.9692} &\textcolor{red}{38.94/0.9694} \\ 
		\cmidrule(r){2-16}
		& Avg. & 26.41/0.7151 & 29.13/0.8052 & 27.10/0.7762 & 28.54/0.7771 & 30.62/0.8311 & 28.42/0.7986 & 28.56/0.7773 & 30.67/0.8305 & 30.74/0.8292 & 30.27/0.8193 & 31.17/0.8428 & 31.26/0.8438 & \textcolor{blue}{31.82/0.8538} &\textcolor{red}{31.93/0.8553} \\ \bottomrule
	\end{tabular}
	\label{Tab:comparison}
\end{table*}


\begin{figure*}[!t]
	\centering
	\includegraphics[width=
   \linewidth]{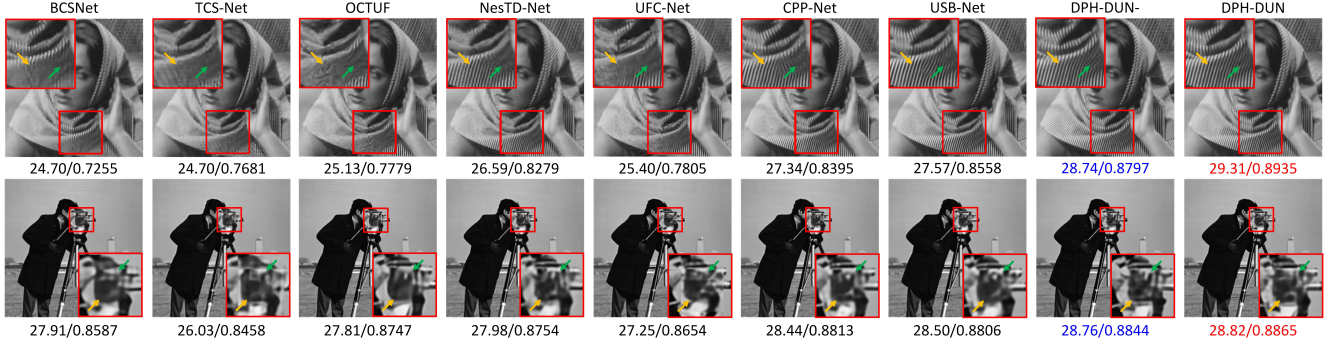}
   \vspace{-0.2in}
	\caption{Visual comparisons on Set11 images, with CS ratio $\gamma=0.10$. The arrows point to details in the image for better comparison.} 
    \vspace{-0.1in}
	\label{figset}
\end{figure*} 
\subsection{Loss Function}
We adopt the standard mean squared error (MSE) as the reconstruction loss. Given the ground-truth images $\{\mathbf{x}_\text{gt}^{j}\}_{j=1}^{N}$ and the network outputs $\{\hat{\mathbf{x}}^{j}\}_{j=1}^{N}$, the loss is formulated as:
\begin{equation}
	\mathcal{L}(\Theta) = \frac{1}{N}\sum\limits_{j=1}^{N} \left\|\hat{\mathbf{x}}^{j} - \mathbf{x}_\text{gt}^{j} \right\|_2^2,
\end{equation}
where $\Theta$ denotes the set of learnable parameters for DPH-DUN, $N$ represents the number of training images. 

\section{Experiments}
\subsection{Experiment Setting}
Our method is evaluated on Urban100 \cite{huang2015single}, Set11 \cite{kulkarni2016reconnet}, Set14 \cite{zeyde2010single} and General100 \cite{dong2016accelerating}, where all images are converted to the Y channel of the YCbCr color space. 
During training, we extract image patches of size $128 \times 128$ with a sensing block size of $32 \times 32$. 
DPH-DUN$^*$ and DPH-DUN denote the lightweight and standard versions, corresponding to 32-channel and 64-channel feature representations, respectively.
The model is trained for 600 epochs on the Waterloo Exploration Database (WED) \cite{7752930} using random cropping and standard data augmentation. The Adam optimizer is employed with a learning rate of $1\times10^{-4}$, and the stage number $K=10$. Experiments are conducted on a single ‌NVIDIA A100 using PyTorch 1.12.

\begin{figure}[b]
  \centering
   \includegraphics[width=\linewidth]{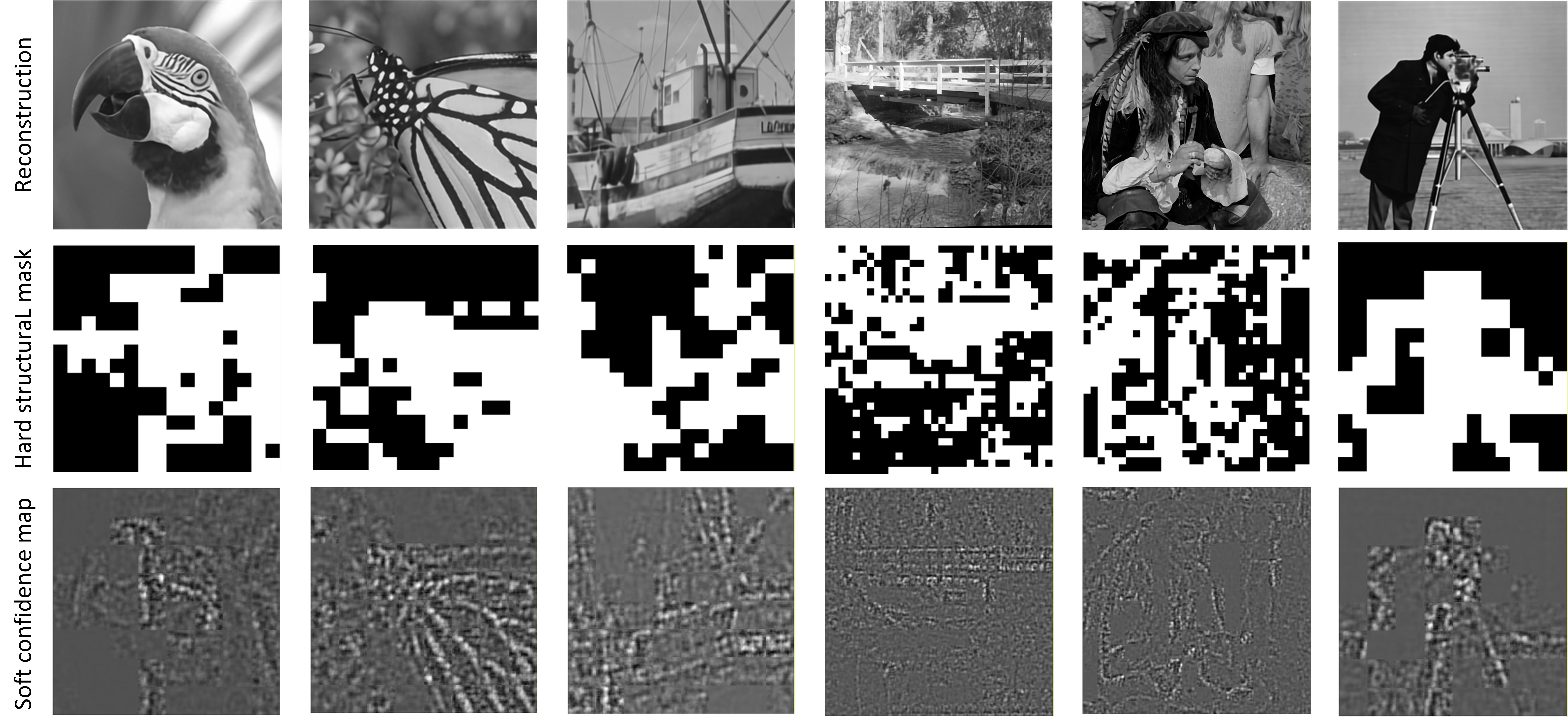}
   \caption{Visualization of the reconstructed image, hard structural mask, and soft confidence map on Set11 and Set14. }
   \vspace{-0.20in}
   \label{fig:mask}
\end{figure}

\subsection{Comparison with Other Methods}
We compare the proposed DPH-DUN with three deep black-box CS methods (BCSNet~\cite{9159912}, CSNet+~\cite{2019Image}, and TCS-Net~\cite{10049603}) and nine deep unfolding CS methods (ISTA-Net+~\cite{zhang2018ista}, AMP-Net~\cite{9298950}, FOSINet~\cite{2022FSOINet}, DPC-DUN~\cite{song2023DPCDUN}), OCTUF~\cite{10205478}, NesTD-Net~\cite{10460452}, UFC-Net~\cite{wang2024ufc}, CPP-Net~\cite{10655712}, and USB-Net~\cite{guo2025usb}) at four different CS ratios. 

As shown in Tab.~\ref{Tab:comparison}, DPH-DUN- and DPH-DUN achieve superior performance over SOTA methods on four benchmark datasets. Compared with the best deep unfolding baseline USB-Net, DPH-DUN gains +1.08dB/+0.0279 on Urban100, +0.53dB/+0.0147 on Set11, +0.55dB/+0.0136 on Set14, and +0.67dB/+0.0115 on General100. It further exceeds the strongest deep black-box baseline BCSNet by +3.79dB/+0.0946, +2.53dB/+0.0696, +2.30dB/+0.0507, and +3.39dB/+0.0782, respectively.

Qualitative comparisons in Fig.~\ref{figset} illustrate that the DPH-DUN better restores complex details, while competing methods often suffer from blurring or artifact amplification. Furthermore, more comparisons can be found in the \textit{Supplementary Material} (\textit{Sec. 3.1}).

\subsection{Ablation Studies and Discussion}
\paragraph{Visualization of Masks and Maps.}
We visualize the hard structural masks and soft confidence maps in Fig.~\ref{fig:mask}. In hard structural masks, brighter regions indicate blocks emphasized by DPH-DUN. In soft confidence maps, brighter pixels are amplified more for feature enhancement. 
This shows that DPH-DUN adaptively focuses on informative regions to improve reconstruction quality.
\vspace{-0.10in}
\paragraph{Different components.}
To assess the contribution of each key component, we conduct ablation experiments on Set11 at CS ratio $\gamma=0.10$. As shown in Tab.~\ref{table:ablation}, both the signal- and gradient-based hyperpriors are crucial for achieving high-quality reconstruction. Removing signal-based hyperpriors or gradient-based hyperpriors leads to notable degradation, highlighting their effectiveness in adaptive optimization and structural perception. In addition, excluding the HHA or HSA weakens regional adaptivity and hierarchical enhancement. Overall, the full DPH-DUN achieves the best performance, demonstrating the complementary benefits of all components.

\begin{figure}[b]
  \centering
  \vspace{-0.1in}
   \includegraphics[width=\linewidth]{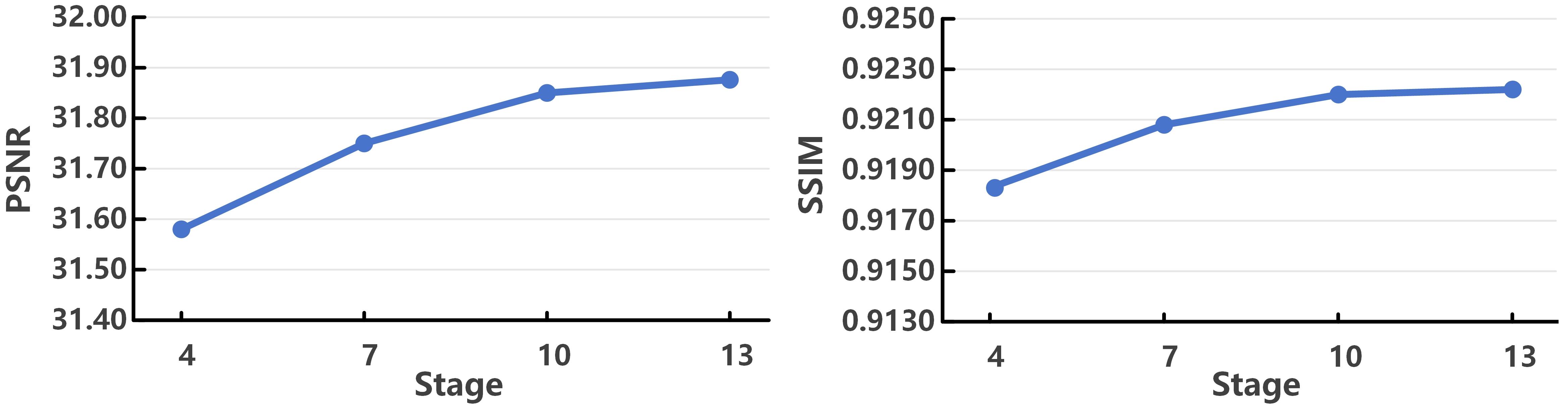}
   \caption{Comparisons of PSNR/SSIM performance on Set11 at
CS ratio $\gamma=0.10$ under different number of stages \(K\). }
   \label{fig:stage}
   \vspace{-0.15in}
\end{figure}

\vspace{-0.1in}
\paragraph{Ratios of Sampling and Top-K.} 
We study the effect of measurement allocation and Top-K selection on Set11 at CS ratio $\gamma=0.10$. Results are shown in Tab.~\ref{table:ratio_ablation}. A 1:4 sampling ratio and $K=50\%$ provide effective reconstruction and structural guidance.
\vspace{-0.1in}
\paragraph{Number of stages.}
To study the impact of stage number, experiments are conducted with $K \in [4,7,10,13]$ on Set11 at CS ratio $\gamma=0.10$. Fig.~\ref{fig:stage} shows that performance improves with K, but the gains saturate at $K = 10$, indicating a trade-off between accuracy and computation.
\vspace{-0.1in}

\begin{table}[t]
\centering
\caption{Ablation on different components on DPH-DUN.}
\vspace{-0.05in}
\footnotesize
\setlength{\tabcolsep}{5.0mm}
\renewcommand{\arraystretch}{0.9}
\label{table:ablation}
\begin{tabular}{l| c}
\toprule
\rowcolor{pink!20}
Cases & PSNR / SSIM \\
\midrule
w/o signal-based hyperpriors (w/o HSSG) & 31.76/0.9201 \\
w/o hard structural mask & 31.74/0.9198 \\
w/o soft confidence map & 31.70/0.9195 \\
w/o gradient-based hyperpriors & 31.66/0.9191 \\
w/o HHA & 31.69/0.9185 \\
w/o HSA & 31.45/0.9140 \\
Full (ours) & 31.86/0.9222 \\
\bottomrule
\end{tabular}
\end{table}


     

\begin{table}[t]
\centering
\caption{Ablation on different ratios of sampling and Top-K.}
\vspace{-0.05in}
\footnotesize
\setlength{\tabcolsep}{5mm}
\renewcommand{\arraystretch}{0.9}
\label{table:ratio_ablation}
\begin{tabular}{c|c|c|c}
\toprule
\rowcolor{pink!20}
Ratio&Sampling& Ratio& Top-K \\
\midrule
1:1  & 31.78/0.9200 & 10\%  & 31.71/0.9189 \\
1:2  & 31.82/0.9218 & 30\%  & 31.79/0.9210 \\
1:4  & 31.86/0.9222 & 50\%  & 31.86/0.9222 \\
1:8  & 31.69/0.9195 & 70\%  & 31.83/0.9219 \\
\bottomrule
\end{tabular}
\vspace{-0.12in}
\end{table}

\begin{figure}[t]
  \centering
   \includegraphics[width=\linewidth]{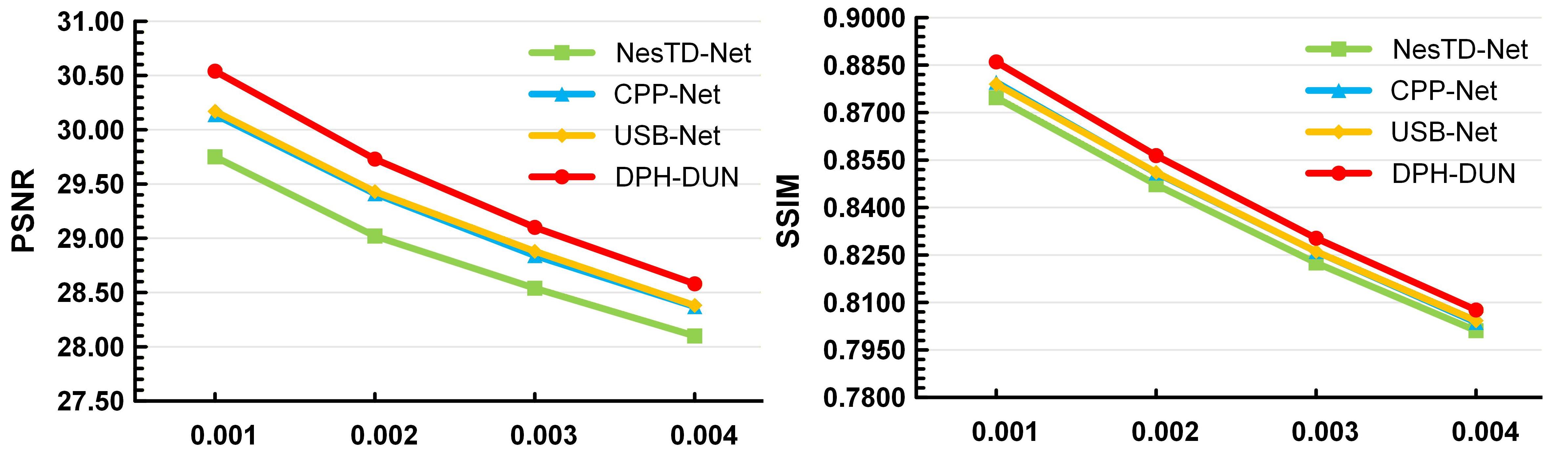}
   \vspace{-0.2in}
   \caption{Comparisons of PSNR/SSIM performance on Set11 at CS ratios $\gamma=0.10$ under different Gaussian noise levels.}
   \label{fig:NOISE}
   \vspace{-0.05in}
\end{figure}

\paragraph{Performance under Noise.}
To evaluate the performance of our method under noise, Gaussian noise with $\sigma\in\left \{0.001,0.002,0.003,0.004\right \}$ is added to Set11. At CS ratio $\gamma=0.10$, our method consistently outperforms baselines in PSNR and SSIM (Fig.~\ref{fig:NOISE}), demonstrating strong noise resilience. Besides, the performance at other CS ratios is presented in the \textit{Supplementary Material} (\textit{Sec. 3.2}).


\subsection{Extension to CS-MRI}
\begin{figure}[t]
  \centering
   \includegraphics[width=\linewidth]{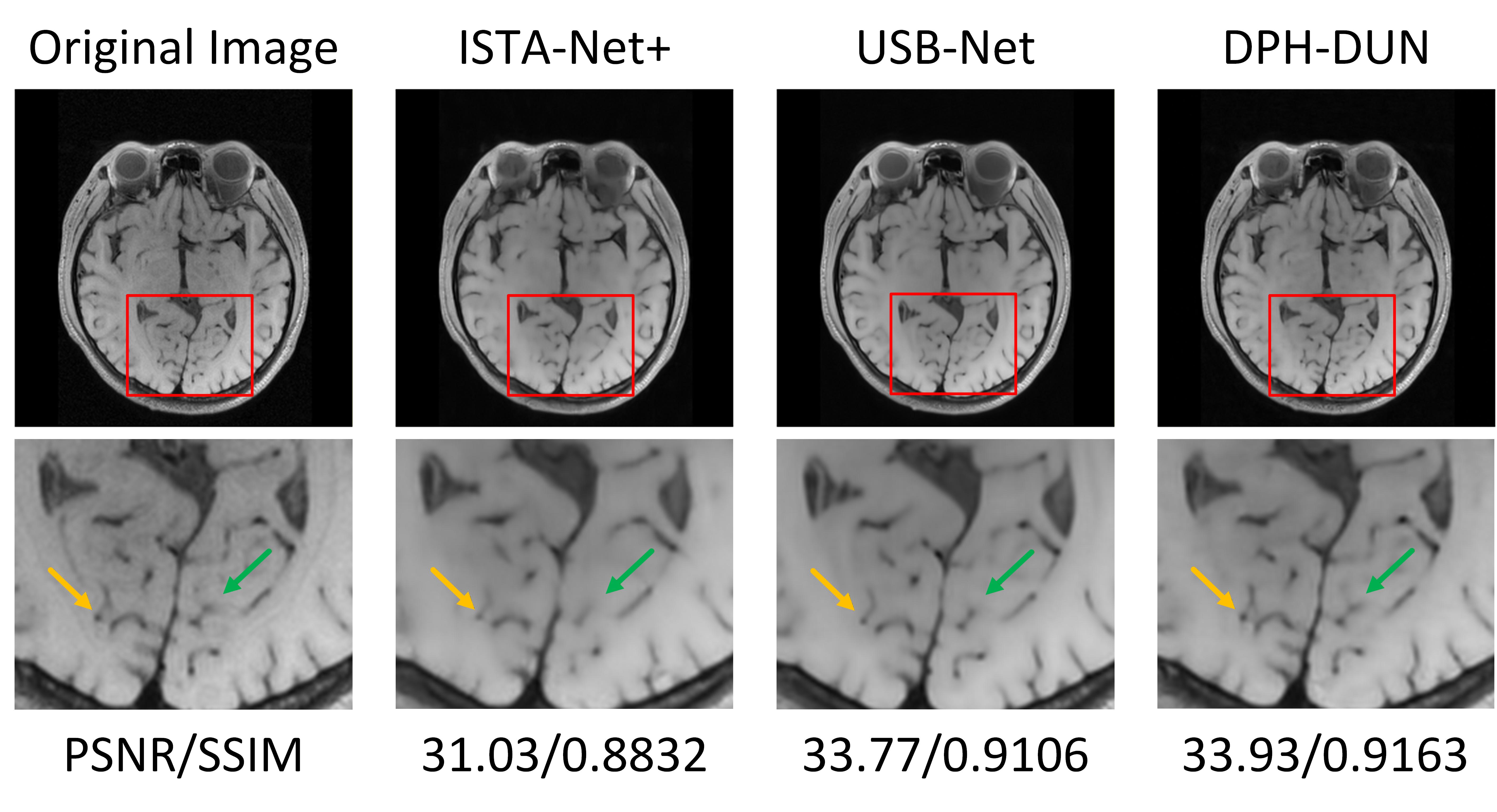}
   \vspace{-0.25in}
   \caption{Visual comparisons on Brain dataset using Pseudo Radial masks at CS ratio $\gamma=0.10$. }
   \label{fig:mri}
   \vspace{-0.1in}
\end{figure}

DPH-DUN is extended to CS-MRI to reconstruct images from undersampled Fourier data. The sampling matrix is defined as $\boldsymbol{\Phi} = \mathbf{SF}$, where $\mathbf{S}$ is a sub-sampling mask and $\mathbf{F}$ denotes the discrete Fourier transform. Using the same brain datasets and pseudo-radial masks as in prior studies \cite{zhang2018ista}, DPH-DUN achieves consistently higher reconstruction quality across all CS ratios, as illustrated in Fig.~\ref{fig:mri} and reported in Tab.~\ref{tablemri}, demonstrating its ability to recover finer structural details compared with competing methods.

\begin{table}[t]
	\centering
    \caption{The performance comparisons of recent CS-MRI methods on Brain dataset at different CS ratios.}
    \vspace{-0.1in}
    \footnotesize
	\setlength{\tabcolsep}{0.55mm}
    \renewcommand{\arraystretch}{0.9}
	\begin{tabular}{c|c|c|c|c}
		\toprule
        \rowcolor{pink!20}
		Methods & 0.05 & 0.10 & 0.20 & Avg. \\
		\midrule
		Zero-filled  & 24.20/0.5417 & 26.81/0.6030 & 30.41/0.7229 & 27.14/0.6225 \\ 
        DC-CNN~\cite{schlemper2017deep}&30.81/0.8370 &34.33/0.8957 &38.43/0.9467 &34.52/0.8931\\
        RDN~\cite{sun2018compressed}&30.95/0.8421 &34.38/0.8998& 38.47/0.9474& 34.60/0.8964\\
        ISTA-Net+~\cite{zhang2018ista}&31.28/0.8547 &34.62/0.9035 &38.57/0.9478 &34.82/0.9020\\
        CDNN~\cite{zheng2019cascaded}& 31.58/0.8513& 34.67/0.9014 &38.65/0.9476& 34.97/0.9001\\
        HiTDUN~\cite{9763318} &32.72/0.8770 &35.71/0.9179& 39.27/0.9529& 35.90/0.9159\\
        USB-Net~\cite{guo2025usb}&  \textcolor{blue}{33.91/0.9014}&\textcolor{blue}{36.29/0.9275}&\textcolor{blue}{39.52/0.9549}&\textcolor{blue}{36.57/0.9279}\\
        \midrule
         DPH-DUN & \textcolor{red}{34.02/0.9049}& \textcolor{red}{36.40/0.9302} & \textcolor{red}{39.65/0.9595} & \textcolor{red}{36.69/0.9315}\\
        \bottomrule
	\end{tabular}
	\label{tablemri}
 \vspace{-0.1in}
\end{table}

\section{Conclusion}
In this paper, a novel Dual-Path Hyperprior Informed Deep Unfolding Network (DPH-DUN) is proposed, leveraging two measurement subsets to enable hyperprior informed reconstruction. Specifically, the Deep Hyperprior Learning branch generates signal-based and gradient-based hyperpriors via lightweight neural modules, providing collaborative guidance for subsequent reconstruction. In the Hyperprior Informed Reconstruction branch, a Hyperprior Informed Step-size Generation network adaptively modulates gradient descent, while two hyperprior informed attention mechanisms selectively refine features in challenging regions. Extensive experiments on benchmark datasets demonstrate that DPH-DUN consistently outperforms existing methods across diverse datasets and various CS ratios.\hspace*{0pt}


{
    \small
    \bibliographystyle{ieeenat_fullname}
    \bibliography{main}
}

\end{document}